\documentclass[conference]{IEEEtran}
\IEEEoverridecommandlockouts
\usepackage{cite}

\usepackage{booktabs} 
\usepackage{bbm}
\usepackage{algorithm}
\usepackage{algorithmic}
\usepackage{amsmath}
\usepackage{amssymb}
\usepackage{graphicx}
\usepackage[utf8]{inputenc} 
\usepackage[tight,footnotesize]{subfigure}
\usepackage{url}
\usepackage{bm}
\usepackage{nicefrac}
\usepackage{subfigure}
\usepackage{caption}
\usepackage{enumitem}
\usepackage{multirow}
\usepackage{color}
\usepackage{float}

\usepackage{array}
\newcolumntype{L}[1]{>{\raggedright\let\newline\\\arraybackslash\hspace{0pt}}m{#1}}
\newcolumntype{C}[1]{>{\centering\let\newline  \\\arraybackslash\hspace{0pt}}m{#1}}
\newcolumntype{R}[1]{>{\raggedleft\let\newline \\\arraybackslash\hspace{0pt}}m{#1}}

\usepackage{pdftexcmds}
\usepackage{catchfile}
\usepackage{ifluatex}
\usepackage{ifplatform}

\ifwindows

\usepackage{times}

\fi
\iflinux
\fi
\ifmacosx
\fi 

\usepackage{array,ragged2e}

\newcolumntype{P}[1]{>{\RaggedRight\hspace{0pt}}p{#1}}

\graphicspath{ {figures/} }

\DeclareMathOperator*{\argmin}{arg\,min}
\DeclareMathOperator*{\argmax}{arg\,max}

\def\be{\textbf{e}}

\def\bH{\textbf{H}}

\def\bW{\textbf{W}}
\def\bX{\textbf{X}}
\def\ba{\textbf{a}}
\def\bh{\textbf{h}}

\def\bw{\textbf{w}}

\def\bz{\textbf{z}}

\def\bal{\bm{\alpha}}

\renewcommand{\arraystretch}{1.2}

\begin{document}

\title{Search to aggregate neighborhood \\ for graph neural network}

\author{\IEEEauthorblockN{Huan ZHAO$^1$, Quanming YAO$^{1,2}$ and Weiwei TU$^1$}
\IEEEauthorblockA{
	\textit{$^1$4Paradigm Inc.} \textit{$^2$Department of Electronic Engineering, Tsinghua University}\\
	\{zhaohuan,yaoquanming,tuweiwei\}@4paradigm.com}
}
\maketitle

\begin{abstract}
Recent years have witnessed the popularity and success of graph neural networks (GNN) in various scenarios. 
To obtain data-specific GNN architectures, researchers turn to neural architecture search (NAS), which have made impressive success in discovering effective architectures in convolutional neural networks. 
However, it is non-trivial to apply NAS approaches to GNN due to challenges in search space design and expensive searching cost of existing NAS methods.
In this work, to obtain the data-specific GNN architectures and address the computational challenges facing by NAS approaches, 
we propose a framework, which tries to Search to Aggregate NEighborhood (SANE), to automatically design data-specific GNN architectures. 
By designing a novel and expressive search space, we propose a differentiable search algorithm, which is more efficient than previous reinforcement learning based methods.
Experimental results on four tasks and seven real-world datasets demonstrate the superiority of SANE compared to existing GNN models and NAS approaches in terms of effectiveness and efficiency. 
\footnote{Code is available at: \url{https://github.com/AutoML-4Paradigm/SANE}.
	\\Correspondence is to Q.Yao.}
\end{abstract}


\section{Introduction}

\begin{figure*}[ht]
	\centering
	\includegraphics[width=0.75\textwidth]{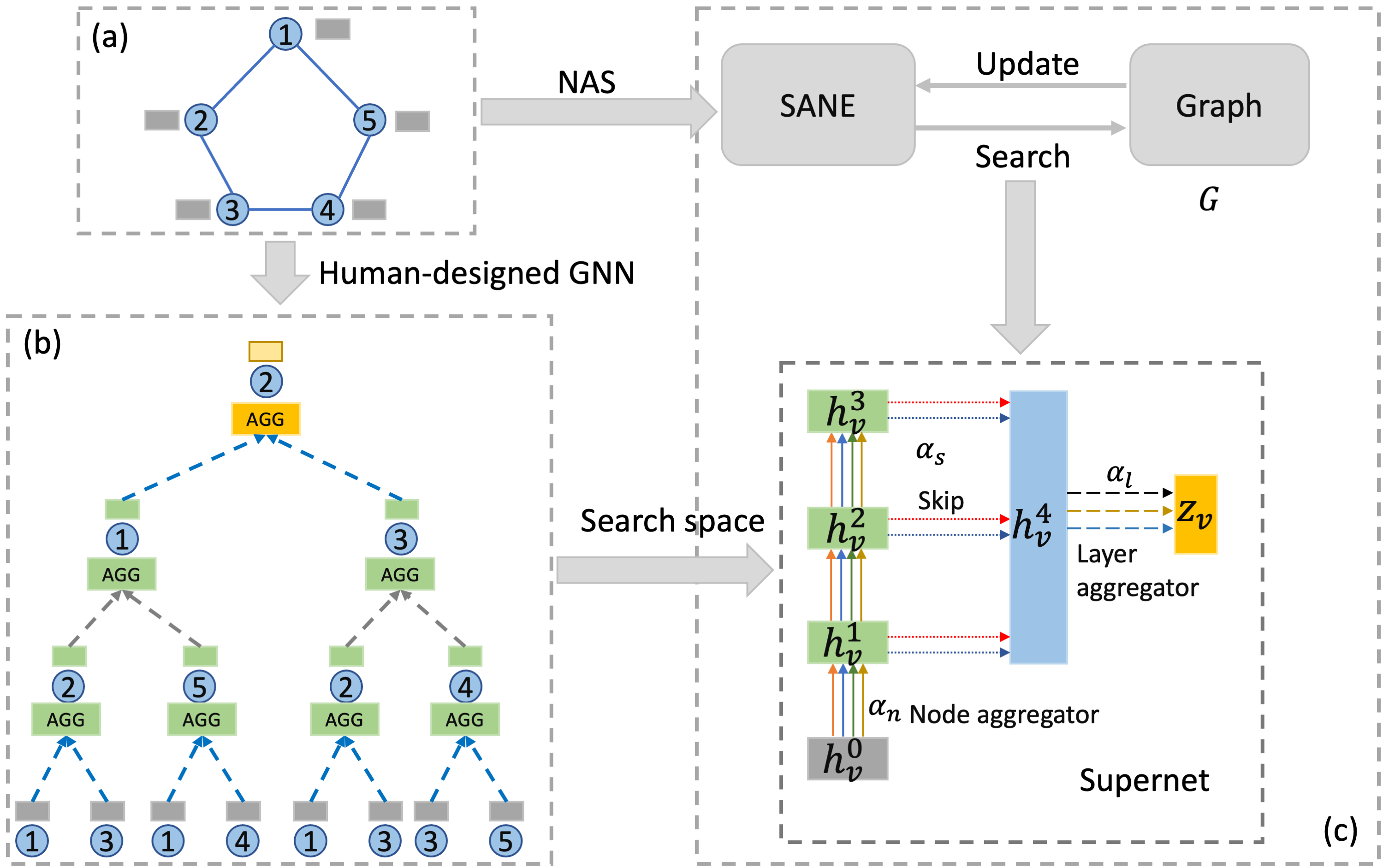}
	\caption{An illustration of the proposed framework. (Best viewed in color)
	(a) Upper Left: an example graph with five nodes. The gray rectangle represents the input features of each node;
	(b) Bottom Left: a typical 3-layer GNN model following the message passing neighborhood aggregation schema, which computes the embeddings of node ``2''; 
	(c) The supernet for a 3-layer GNN, introduced in Section~\ref{sec-framework-search-algorithm}, and $\bal_n,\bal_l, \bal_s$ represent, respectively, weight vectors for node aggregators, layer aggregators, and skip-connections in the corresponding edges. 
	The rectangles denote the representations, out of which three green ones represent the hidden embeddings, gray ($\bh_v^0$) and yellow ($\bz_v$) ones represent the input and output embeddings, respectively, and blue one ($\bh^4_v$) represents the set of output embeddings of three node aggregators for the layer aggregator.}
	\label{fig-sane-framework}
\end{figure*}

Graph neural networks (GNNs)~\cite{battaglia2018relational,gori2005new} 
have been extensively researched in the past five years, and show promising results on various graph-based tasks, 
e.g., node classification~\cite{hamilton2017inductive,velivckovic2017graph,wang2019heterogeneous}, 
recommendation~\cite{ying2018graph,wang2018billion,xiao2019beyond,li2020hierarchical,zheng2020price}, 
fraud detection~\cite{zhang2020fakedetector}, chemistry~\cite{gilmer2017neural} and travel data analysis~\cite{yang2018did}. 
In the literature, various GNN architectures~\cite{kipf2016semi,hamilton2017inductive,velivckovic2017graph,gao2018large,xu2018powerful,xu2018representation,liu2019geniepath} have been designed for different tasks, and most of these approaches are relying on a neighborhood aggregation
(or \textit{message passing}) schema~\cite{gilmer2017neural} (see the example in Figure~\ref{fig-sane-framework}(a) and (b)).
Despite the success of GNN models, they are facing a major challenge. That is there is no single model that can perform the best on all tasks and no optimal model on different datasets even for the same task (see the experimental results in Table~\ref{tb-performance-trans-induc}). 
Thus given a new task or dataset, huge computational and expertise resources would be invested to find a good GNN architecture, 
which limits the application of GNN models. Moreover, existing GNN models do not make full use of the best architecture design practice. For example, existing GNN models tend to stack multiple layers with the same aggregation function to aggregate hidden features of neighbors, however, it remains to be seen whether different combinations of aggregation functions can further improve the performance. In one word, it is left to be explored whether we can obtain data-specific GNN architectures beyond existing ones. This problem is dubbed \textit{architecture challenge}.

To address this architecture challenge, researchers turn to neural architecture search (NAS)~\cite{zoph2016neural,baker2016designing}, which has been a hot topic since it shows promising results in automatically designing novel and better neural architectures beyond human-designed ones. 
For example, in computer vision, the searched architectures by NAS can beat the state-of-the-art human-designed ones on CIFAR-10 and ImageNet datasets by a large margin~\cite{cai2018proxylessnas,tan2019efficientnet}.
Motivated by such a success, very recently, two preliminary works, GraphNAS~\cite{gao2019graphnas} and Auto-GNN~\cite{zhou2019auto}, made the first attempt to tackle the architecture challenge in GNN with the NAS approaches.
However, it is non-trivial to apply NAS to GNN. 
One of the key components of NAS approaches~\cite{elsken2018neural,Bender18one-shot} is the design of search space, 
i.e., defining what to search, which directly affects the effectiveness and efficiency of the search algorithms. 
A natural search space is to include
 all hyper-parameters related to GNN models, e.g., the hidden embedding size, aggregation functions, and number of layers, as done in GraphNAS and Auto-GNN. 
 However, this straightforward design of search space in GraphNAS and Auto-GNN have two problems. The first one is that various GNN architectures, e.g., GeniePath~\cite{liu2019geniepath}, or JK-Network~\cite{xu2018representation}, are not included, thus the best performance might not be that good. 
The second one is that it makes the architecture search process too expensive in GraphNAS/Auto-GNN by incorporating too many hyper-parameters into the search space. In NAS literature, it remains a challenging problem to design a proper search space, which should be expressive (large) and compact (small) enough, thus a good balance between accuracy and efficiency can be achieved.

Besides, existing NAS approaches for GNNs are facing an inherent challenge, which is that they are extremely expensive due to the trial-and-error nature, 
i.e., one has to train from scratch and evaluate as many as possible candidate architectures over the search space before obtaining a good one~\cite{zoph2016neural,liu2018darts,elsken2018neural}. 
Even in small graphs, in which most existing human-designed GNN architectures are tuned, 
the search cost of NAS approaches, i.e., GraphNAS and Auto-GNN, can be quite expensive. 
This challenge is dubbed \textit{computational challenge}.

In this work,  to address the architecture and computational challenges, 
we propose a novel NAS framework, 
which tries to Search to Aggregate NEighborhood (SANE) for automatic architecture search in GNN. 
By revisiting extensive GNN models, 
we define a novel and expressive search space, 
which can emulate more human-designed GNN architectures than existing NAS approaches, i.e., GraphNAS and Auto-GNN. 
To accelerate the search process, 
we adopt the advanced one-shot NAS paradigm~\cite{liu2018darts}, and design a differentiable search algorithm, which trains a~\textit{supernet} subsuming all candidate architectures, thus greatly reducing the computational cost. 
We further conduct extensive experiments on three types of tasks, including transductive, inductive, and database (DB) tasks, to demonstrate the effectiveness and efficiency of the proposed framework.
To summarize, the contributions of this work are in the following: 
\begin{itemize}[leftmargin=*]
\item In this work, to address the \textit{architecture challenge} in GNN models, we propose the SANE framework based on NAS. By designing a novel and expressive search space, SANE can emulate more human-designed GNN architectures than existing NAS approaches.

\item To address the \textit{computational challenge}, we propose a differentiable architecture search algorithm, which is efficient in nature comparing to the trial-and-error based NAS approaches. To the best of our knowledge, this is the first differentiable NAS approach for GNN.


\item Extensive experiments on five real-world datasets are conducted to compare SANE with human-designed GNN models and NAS approaches. 
The experimental results demonstrate the superiority of SANE in terms of effectiveness and efficiency compared to all baseline methods.
\end{itemize}


\noindent\textbf{Notations.}
Formally, let $\mathcal{G} = (\mathcal{V}, \mathcal{E})$ be a simple graph with node features 
$\bX \in \mathbb{R}^{N \times d}$, 
where $\mathcal{V}$ and $\mathcal{E}$ represent the node and edge sets, respectively. $N$ represents the number of the nodes, and $d$ is the dimension of node features, 
We use $N(v)$ to represent the first-order neighbors of a node $v$ in $\mathcal{G}$, i.e., $N(v) = \{u \in \mathcal{V}|(v,u) \in \mathcal{E}\}$. In the literature, we further create a new set $\widetilde{N}(v)$, which is the neighbor set including itself, 
i.e.,  $\widetilde{N}(v) =  \{v\} \cup \{u \in \mathcal{V}|(v,u) \in \mathcal{E}\} $. A new graph $\mathcal{\widetilde{G}}$ is always created by adding self-loop to every $v \in \mathcal{V}$.

\section{Related Works}
\label{sec-rel}


\subsection{Graph Neural Network (GNN)}
\label{sec-rel-gnn}

GNN was first proposed in~\cite{gori2005new}
and many of its variants~\cite{kipf2016semi,hamilton2017inductive,velivckovic2017graph,gao2018large,xu2018powerful,xu2018representation,liu2019geniepath} have been proposed in the past five years. 
Generally, these GNN models can be unified by a neighborhood aggregation or message passing schema~\cite{gilmer2017neural}, 
where the representation of each node is learned by iteratively aggregating the embeddings (``message'') of its neighbors. 
A typical $K$-layer GNN in the neighborhood aggregation schema can be written as follows (see the illustrative example in Figure~\ref{fig-sane-framework}(a) and (b)): the $l$-th layer
$(l = 1,\cdots, k)$ updates $\bh_v$ for each node $v$ as
\begin{align}
\label{eq-mpnn}
\bh_v^l =  \sigma(\bW^{l} \cdot \text{AGG}_{\text{node}}(\{\bh_u^{l-1}, \forall u \in \widetilde{N}(v)\})),
\end{align}
where 
$\bh_v^l \in \mathbb{R}^{d_l}$ represents the hidden features of a node $v$ learned by the $l$-th layer, and $d_l$ is the corresponding dimension. $\bW^{l}$ is a trainable weight matrix shared by all nodes in the graph, and $\sigma$ is a non-linear activation function, e.g., sigmoid or ReLU. $\text{AGG}_{\text{node}}$ is the key component, i.e., a pre-defined aggregation function, which varies across different GNN models. 
For example, a weighted summation function is designed as $\text{AGG}_{\text{node}}$ in~\cite{kipf2016semi}, 
and different functions, e.g., mean and max-pooling, are proposed as the node  aggregators in~\cite{hamilton2017inductive}. 
Further, to weigh the importance of different neighbors, 
attention mechanism is incorporated to design the node aggregators~\cite{velivckovic2017graph}.
For more details of different node aggregators,  we refer readers to Table~\ref{tb-node-agg-detail} in Appendix~\ref{append-node-agg}.

Motivated by the success of residual network~\cite{he2016deep},
residual mechanisms are incorporated to improve the performance of GNN models. In \cite{chen2020simple}, two simple residual connections are designed to improve the performance of the vanilla GCN model. And in \cite{xu2018representation}, skip-connections are used to propagate message from intermediate layers to the last layer, then the final representation of the node $v$ is computed by a layer aggregator as 
\begin{align*}
\bz_v  =  \text{AGG}_{\text{layer}}( \bh_v^{1},\cdots, \bh_v^{K} ),
\end{align*} 
where $\text{AGG}_{\text{layer}}$ can be different operations, e.g., max-pooling, concatenation. 
In this way, neighbors in long ranges are used together to learn the final representation of each node, 
and the performance gain is reported with the layer aggregator~\cite{xu2018representation}.
With the node and layer aggregators, we point out the two key components of exiting GNN models, i.e., the neighborhood aggregation function and the range of the neighborhood, which are tuned depending on the tasks~\cite{xu2018representation}. In Section~\ref{sec-framework-search-space}, we will introduce the search space of SANE based on these two components.

\subsection{Neural Architecture Search (NAS)}
\label{sec:nas}

Neural architecture search (NAS) \cite{baker2016designing,zoph2016neural,elsken2018neural,yao2018taking} aims to automatically find unseen and better architectures comparing to expert-designed ones, which have shown promising results in searching for convolutional neural networks (CNN)~\cite{baker2016designing,zoph2016neural,elsken2018neural,liu2018darts,zoph2018learning,tan2019efficientnet,yao2019differentiable}. 
Early NAS approaches follow a trial-and-error pipeline, 
which firstly samples a candidate architecture from a pre-defined search space, then trains it from scratch, and finally gets the validation accuracy. This process is repeated many times before obtaining an architecture with satisfying performance.
Representative methods are reinforcement learning (RL) algorithms \cite{baker2016designing,zoph2016neural}, 
which are inherently time-consuming to train thousands of candidate architectures during the search process. To address the efficiency problem, a series of methods adopt weight sharing strategy to reduce the computational cost.
To be specific, instead of training one by one thousands of separate models from scratch, 
one can train a single large network (\textit{supernet}) capable of emulating any architecture in the search space. 
Then each architecture can inherit the weights from the supernet, and the best architecture can be obtained more efficiently. 
This paradigm is also referred to as ``one-shot NAS'', and representative methods are~\cite{pham2018efficient,Bender18one-shot,liu2018darts,cai2018proxylessnas,yao2019differentiable,xie2018snas,zhou2019bayesnas,akimoto2019adaptive}.

Recently, to obtain data-specific GNN architectures, several works based on NAS were proposed.
GraphNAS~\cite{gao2019graphnas} and Auto-GNN\cite{zhou2019auto} made the first attempt to introduce NAS into GNN.
In~\cite{peng2019learning}, an evolutionary-based search algorithm is proposed to search architectures on top of the Graph Convolutional Network (GCN)~\cite{kipf2016semi} model, and action recognition problem is considered. However, in this work, we focus on the node representation learning problem, 
which is an important one in GNN literature. In \cite{lai2020policy}, a RL-based method is proposed to search for node-specific layer numbers given a GNN model, e.g., GCN or GAT, and in \cite{ding2020propagation}, propagation matrices in the message passing framework are searched. These two works can be regarded as orthogonal works of our framework.
For GraphNAS and Auto-GNN, they are RL-based methods, thus very expensive in nature. Besides, the search space of the proposed SANE is more expressive than those of GraphNAS and Auto-GNN. In \cite{zhao2020simplifying}, the search space of GraphNAS is further simplified by introducing node and layer aggregators. 
However, the search method is still RL-based one.
Further, to address the computational challenges, we design a differentiable search algorithm based on the one-shot method~\cite{liu2018darts}. 
To the best of our knowledge, this is the first differentiable NAS approach for architecture search in GNN.

\section{The Proposed Framework}
\label{sec-framework}

\subsection{The Search Space Design}
\label{sec-framework-search-space}

\begin{table}[t]
	\caption{The operations we use as node and layer aggregators for the search space of SANE.}
	\label{tb-search-space-ops}
	\setlength\tabcolsep{5pt}
	\centering
	\begin{tabular}{c|C{200px}}
		\toprule
		&                                     Operations                                      \\ \midrule
		$\mathcal{O}_n$ &    \texttt{SAGE-SUM}, \texttt{SAGE-MEAN}, \texttt{SAGE-MAX},    
		\texttt{GCN}, \texttt{GAT},\texttt{GAT-SYM},  \texttt{GAT-COS}, \texttt{GAT-LINEAR}, 
		\texttt{GAT-GEN-LINEAR}, \texttt{GIN}, \texttt{GeniePath}       \\ \midrule
		$\mathcal{O}_l$         &      \texttt{CONCAT}, \texttt{MAX}, \texttt{LSTM}       \\ \midrule
		$\mathcal{O}_s$          &                          \texttt{IDENTITY}, \texttt{ZERO}                           \\ \bottomrule
	\end{tabular}
\end{table}

\begin{table*}[t]
	\caption{Comparisons between existing GNN models and the proposed SANE. Note that the variants of attention is from~\cite{gao2019graphnas}, which represents different aggregation methods despite that they all use the attention mechanism.}
	\label{tb-gnn-comparison}
	\centering
	\begin{tabular}{C{60px} | c | c | c | c}
		\toprule
		               & Model                                  & Node aggregators                                                                    & Layer aggregators & Emulate by SANE \\ \midrule
		               & GCN~\cite{kipf2016semi}                & \texttt{GCN}                                                                        & $\times$          & $\checkmark$    \\ \cmidrule{2-5}
		               & SAGE~\cite{hamilton2017inductive}      & \texttt{SAGE-SUM}/\texttt{-MEAN}/\texttt{-MAX}                       & $\times$          & $\checkmark$    \\ \cmidrule{2-5}
		               & GAT~\cite{velivckovic2017graph}        & \texttt{GAT}, \texttt{GAT-SYM}/\texttt{-COS}/ \texttt{-LINEAR}/\texttt{-GEN-LINEAR} & $\times$          & $\checkmark$    \\ \cmidrule{2-5}
		Human-designed & GIN~\cite{xu2018powerful}              & \texttt{GIN}                                                                        & $\times$          & $\checkmark$    \\ \cmidrule{2-5}
		         architectures      & LGCN~\cite{gao2018large}               & \texttt{CNN}                                                                        & $\times$          & $\checkmark$    \\ \cmidrule{2-5}
		               & GeniePath~\cite{liu2019geniepath}      & \texttt{GeniePath}                                                                  & $\times$          & $\checkmark$    \\ \cmidrule{2-5}
		               & JK-Network~\cite{xu2018representation} & depends on the base GNN                                                             & $\checkmark$      & $\checkmark$    \\ \midrule
		     NAS       & SANE                                   & learned combination of aggregators                                                  & $\checkmark$      &                 \\ \bottomrule
	\end{tabular}
\end{table*}

In the literature~\cite{Bender18one-shot,elsken2018neural}, 
designing a good search space is very important for NAS approaches. 
On one hand, a good search space should be  
large and expressive
enough to emulate various existing GNN models, 
thus ensures the competitive performance (see Table~\ref{tb-performance-trans-induc}).
On the other hand, 
the search space should be small and compact enough for the sake of computational resources, i.e., searching time (see Table~\ref{tb-total-cost}).
In the well-established work~\cite{xu2018powerful}, the authors shows that the expressive capability is dependent on the properties of different aggregation functions, thus to design an expressive yet simplified search space, we focus on two key important components:  node and layer aggregators,
which are introduced in the following:
\begin{itemize}[leftmargin=*]
\item \textbf{Node aggregators}: 
We choose 11 node aggregators based on popular GNN models, and they are presented in Table~\ref{tb-search-space-ops}. We denote the node aggregator set by $\mathcal{O}_n$. Note that the detail of each node aggregator is given in Table~\ref{tb-node-agg-detail} in Appendix~\ref{append-node-agg}.
 
\item \textbf{Layer aggregators}: 
We choose 3 layer aggregators as shown in Table~\ref{tb-search-space-ops}. Besides, we have two more operations, \texttt{IDENTITY} and \texttt{ZERO}, related to skip-connections. Instead of requiring skip-connections between all intermediate layers and the last layer in JK-Network, in this work, we generalize this option by proposing to search for the existence of skip-connections between each intermediate layer and the last layer. To connect, we choose \texttt{IDENTITY}, and \texttt{ZERO} otherwise. We denote the layer aggregator set by $\mathcal{O}_l$ and skip operation set by $\mathcal{O}_s$.

\end{itemize}


To show the expressive capability of the designed search space, 
here we further give a detailed comparison between SANE and existing GNN models in Table~\ref{tb-gnn-comparison}, 
from which we can see that SANE can emulate existing models. Besides, we also discuss the connections between SANE and more recent advanced GNN baselines in Appendix \ref{append-baseline}.

\subsection{Differentiable Architecture Search}
\label{sec-framework-search-algorithm}

In this part, we first introduce how to represent the search space of SANE as a supernet, 
which is a directed acyclic graph (DAG) in Figure~\ref{fig-sane-framework}(c), 
and then how to use gradient descent for the architecture search.

\subsubsection{Continuous Relaxation of the Search Space}
\label{sec:das}

Assume we use a $K$-layer GNN with JK-Network as backbone ($K$ = 3 in Figure~\ref{fig-sane-framework}(c)), 
and the supernet has $K+3$ nodes, where each node $x^{l}$ is a latent representation, e.g., the input features of a node, or the embeddings in the intermediate layers. Each directed edge $(i, j)$ is associated with an operation $o^{ij}$ that transforms $x^l$, e.g., GAT aggregator. 
Without loss of generality, we have one input node, one output node, and one node representing the set of all skipped intermediated layers, thus we have $K + 3$ node for the supernet in total. 
Then the task is transformed to find a proper operation on each edge, leading to a discrete search space, 
which is difficult in nature.

Motivated by the differentiable architecture search in~\cite{liu2018darts}, 
we relax the categorical choice of a particular operation to a softmax over all possible operations:
\begin{equation}
\bar{o}^{ij}(x) = \sum_{o \in \mathcal{O}}
\frac{\exp(\bal^{ij}_o)}{\sum_{o' \in \mathcal{O}}\exp(\bal_{o'}^{ij}) }o(x),
\label{eq-continuous-relaxation}
\end{equation}
where the operation mixing weights for a pair of nodes $(i,j)$ are parameterized by a vector $\bal^{ij} \in \mathbb{R}^{|\mathcal{O}|}$, 
and $\mathcal{O}$ is chosen from the three operation sets: $\mathcal{O}_n, \mathcal{O}_l, \mathcal{O}_s$ as introduced in Section~\ref{sec-framework-search-space}. Then we have the corresponding $\bal_n, \bal_l, \bal_s$. $x$ represents the input hidden features for a GNN layer, e.g., $\{\bh_u^{(l-1)}, \forall u \in \widetilde{N}(v)\}$.

Let $\bar{o}_n, \bar{o}_s$ and $\bar{o}_l$ are the mixed operations from $\mathcal{O}_n, \mathcal{O}_s, \mathcal{O}_l$ based on
\eqref{eq-continuous-relaxation}, respectively, and we remove the superscript of $o^{ij}$ for simplicity when there is no misunderstanding.
Then given a node $v$ in the graph, the neighborhood aggregation process by SANE is
\begin{align}
\bh_v^{l} 
=  \sigma( \bW_n^{l} \cdot \bar{o}_n (\{\bh_u^{l-1}, \forall u \in \widetilde{N}(v)\})),
\label{eq-sane-con-formula-1}
\end{align}
where $\bW_n^{l}$ is shared by candidate architectures from the search space by each node aggregator.
Then,
for the last layer for the node $v$, the embeddings can be computed by
\begin{align}
\bH_v^{K+1} 
& = \left[ 
\bar{o}_{s}( \bh_v^1), \cdots, \bar{o}_s(\bh_v^K)
\right],
\label{eq-sane-con-formula-2} 
\\
\bz_v 
& = \bar{o}_l\left( \bH_v^{K+1} \right),
\label{eq-sane-con-formula-3}
\end{align}
where $\left[\cdot\right]$ represents we stack all the embeddings from $K$ intermediate layers for the last layer.
From the above equations, we can see that the computing process is the summation of all operations, i.e., aggregator or skip, from the corresponding set, which is what a ``supernet'' means. After we obtain the final representation of the node $\bz_v$,  we can inject it to different types of loss depending on the given task. 
Thus, SANE is to solve a bi-level optimization problem as:
\begin{align}
\min_{\bal \in \mathcal{A}} & \;
\mathcal{L}_{\text{val}} (\bw^*(\bal), \bal),
\label{eq-nested-nas-opt}
\\\
\text{\;s.t.\;} \bw^*(\bal) 
& = \argmin_\bw \mathcal{L}_{\text{tra}}(\bw, \bal),
\label{eq-nested-nas-opt:2}
\end{align}
where $\mathcal{L}_{\text{tra}}$ and $\mathcal{L}_{\text{val}}$ are the training and validation loss, respectively. 
$\bal = \{\bal_n, \bal_s, \bal_l\}$ represents a network architecture, and $\bw^*(\bal)$ the corresponding weights after training. 
In the experiments, we focus on the node classification task, thus cross-entropy loss is used. 
Thus, the task of architecture search by SANE is to learn three continuous variables $\bal_n, \bal_s, \bal_l$ with each 
$\bal = \{\bal^{ij}\}$. 

%


 \begin{table*}[ht]
 	\centering
 	\caption{A detailed comparison between SANE and \mbox{existing} NAS methods for GNN.}
 	\begin{tabular}{c| c | c | c }
 		\toprule
 		& \multicolumn{2}{c|}{Search space} & \multirow{2}{*}{Search algorithm }  \\ \cmidrule{2-3}
 		& Node aggregators & Layer aggregators &   \\ \midrule
 		GraphNAS, Auto-GNN &   $\surd$           &      $\times$           &            RL          \\ \midrule
 		Policy-GNN &      $\times$          &       $\times$           &            RL           \\ \midrule
 		SANE &    $\surd$             &       $\surd$               &  Differentiable                 \\ \bottomrule
 	\end{tabular}
 	\label{tb-compare-graphnas}
 \end{table*}
 
\subsubsection{Optimization by gradient descent}
We can observe that the SANE problem is a bi-level optimization problem, 
where the architecture parameters ${\bm{\alpha}}$ is optimized on validation data (i.e., \eqref{eq-nested-nas-opt}), 
and the network weights $\bw$ is optimized on training data (i.e., \eqref{eq-nested-nas-opt:2}).  
With the above continuous relaxation, 
the advantage is that the recently developed one-shot NAS approach~\cite{liu2018darts} can be applied.

The optimizing detail is given in Algorithm~\ref{alg-sane}.
Specifically, following \cite{liu2018darts,yao2019differentiable},
we give the gradient-based approximation to update the architecture parameters, i.e., 
\begin{align} \label{eq:grad}
\nabla_{\bm{\alpha}}\mathcal{L}_{\text{val}}
(\bw^*({\bm{\alpha}}), {\bm{\alpha}}) \approx \nabla_{\bm{\alpha}}\mathcal{L}_{\text{val}}(\bw - \xi\nabla_{\bw}\mathcal{L}_{\text{tra}}(\bw,{\bm{\alpha}}), {\bm{\alpha}}),
\end{align}
where $\bw$ is the current weight, 
and $\xi$ is the learning rate for a step of inner optimization in \eqref{eq-nested-nas-opt:2} for $\bw$. 
Thus, instead of obtaining the optimized $\bw^*({\bm{\alpha}})$, 
we only need to approximate it by adapting $\bw$ using only a single training step. 
After training, we retain the top-$k$ strongest operations, i.e, the largest weights according \eqref{eq-continuous-relaxation}, and form the complete architecture with the searched operations. Then the searched architecture is re-trained from scratch and tuned on the validation data to obtain the best hyper-parameters. 
Note that in the experiment, we set $k = 1$ for simplicity, which means that a discrete architecture is obtained by replacing each mixed operation $\bar{o}^{ij}$ with the operation of the largest weight, i.e, $o^{ij} = \argmax_{o \in \mathcal{O}}\bal_{o}^{ij}$. 

\begin{algorithm}[ht]
	\caption{SANE - Search to Aggregate NEighborhood.}
	\begin{algorithmic}[1]
		\REQUIRE {The search space $\mathcal{A}$, the number of top architectures $k$, the epochs $T$ for search.}
		\ENSURE The $k$ searched architectures $\mathcal{A}_k$.
		\WHILE{$t = 1, \cdots, T$}
		\STATE{Compute the validation loss $\mathcal{L}_{\text{val}}$;}
		\STATE{Update $\bal_n$, $\bal_s$ and $\bal_l$ by gradient descend rule \eqref{eq:grad}
			 with \eqref{eq-sane-con-formula-1},
			 \eqref{eq-sane-con-formula-2} 
			 and \eqref{eq-sane-con-formula-3} respectively; }
		\STATE{Compute the training loss $\mathcal{L}_{\text{tra}}$;}
		\STATE{Update weights $\bw$ by descending $\nabla_{\bw}\mathcal{L}_{\text{tra}}(\bw, \bal)$ with 
			the architecture $\bal = \{ \bal_n,  \bal_s,  \bal_l\}$; }
		\ENDWHILE
		\STATE{Derive the final architecture \{$\bal_n^*,  \bal_s^*,  \bal_l^*$\}
			based on the trained \{$\bal_n,  \bal_s,  \bal_l$\};}
		\RETURN Searched \{$\bal_n^*,  \bal_s^*,  \bal_l^*$\}.
	\end{algorithmic}
	\label{alg-sane}
\end{algorithm}

\subsection{Comparison with Existing NAS Methods}
\label{sec-framework-graphnas-comparison}

In this part, we further give a comparison between SANE and GraphNAS \cite{gao2019graphnas}, Auto-GNN \cite{zhou2019auto}, and Policy-GNN \cite{lai2020policy}, which are the latest NAS methods for GNN in node-level representation learning. The results are shown in Table~\ref{tb-compare-graphnas}. 
We can see that the advantages of the expressive search space and differentiable search algorithm are evident. Besides, Policy-GNN focuses on searching for the number of layers given a GNN backbone, which can be regarded as an orthogonal work of SANE, then here we discuss more about the comparisons between GraphNAS/Auto-GNN and SANE.

The key difference is that SANE does not include parameters like hidden embedding size, number of attention heads, which tends to be called hyper-parameters of GNN models. The underlying philosophy is that the expressive capability of GNN models is mainly relying on the properties of the aggregation functions, as shown in the well-established work~\cite{xu2018powerful}, thus we focus on including more node aggregators to guarantee as powerful as possible the searched architectures. The layer aggregators are further included to alleviate the over-smoothing problems in deeper GNN models~\cite{xu2018representation}. 
Moreover, this simplified and compact search space has a side advantage, which is that the search space is made smaller in orders, thus the cost of architecture search is reduced in orders. For example, when considering the search space for a $3$-layer GNN in our experiments,  
the total number of architectures in the search space is $11^3 \times 2^3 \times 3 = 31,944  $. 
While in Auto-GNN, there are $ (14112)^3 \approx 2.8 \times 10^{12}$ candidate architectures to be searched~\cite{zhou2019auto}. Finally, since the hyper-parameters are tuned by retraining the derived GNN architectures by Algorithm~\ref{alg-sane}, which is also a standard practice in CNN architecture search~\cite{liu2018darts,xie2018snas}, SANE actually decouples the architecture search and hyper-parameters tuning, while GraphNAS/Auto-GNN mix them up. In Section~\ref{sec-exp-abl-search-space}, we show that by running GraphNAS over the search space of SANE, the performance can be improved, which means that better architectures can be obtained given the same time budget, thus demonstrating the advantages of the decoupling process as well as the simplified and compact search space.

\section{Experiment}
\label{sec-exp}

In this section, we conduct extensive experiments to demonstrate the superiority of the propose SANE in three tasks: transductive task, inductive task, and DB task.

\subsection{Experimental Settings}
\subsubsection{Datasets and Tasks}
Here, we introduce details of different tasks and corresponding datasets (Table~\ref{tb-datasets}). 
Note that transductive and inductive tasks are standard ones in the literature \cite{kipf2016semi,hamilton2017inductive,xu2018representation}.
We further add one popular database (DB) task, entity alignment in cross-lingual knowledge base (KB), to show the capability of SANE in broader domains.

\begin{table}[t]
	\caption{Dataset statistics of the datasets in the experiments. 
		N, E, F and C denote the number of ``Nodes'', ``Edges'', ``Features'' and ``Classes'', respectively. }
	\centering
	\label{tb-datasets}
	\begin{tabular}{c|c|cccc}
		\toprule
		Task &Dataset         & N  &  E  & F & C \\ \midrule
		\multirow{3}{*}{\begin{tabular}[c]{@{}c@{}}Transductive\end{tabular}}&Cora           & 2,708 & 5,278  & 1,433    &    7    \\
		&CiteSeer           & 3,327 & 4,552  &   3,703    &    6    \\
		&PubMed           & 19,717 & 44,324  &   500    &    3    \\\midrule
		Inductive &PPI           & 56,944 & 818,716  &   121    &   50    \\
		\bottomrule
	\end{tabular}
\end{table}

\noindent \textbf{Transductive Task.}  
Only a subset of nodes in one graph is allowed to access as training data, and other nodes are used as validation and test data. 
For this setting, we use three benchmark datasets: Cora, CiteSeer, PubMed. 
They are all citation networks, provided by~\cite{sen2008collective}. 
Each node represents a paper, and each edge represents the citation relation between two papers. 
The dataset contains bag-of-words features for each paper (node), and the task is to classify papers into different subjects based on the citation networks.
We split the nodes in all graphs into 60\%, 20\%, 20\% for training, validation, and test.

\noindent \textbf{Inductive Task.}  
In this task, we use a number of graphs as training data, and other completely unseen graphs as validation/test data. 
For this setting, we use the PPI dataset, provided by~\cite{hamilton2017inductive}, on which the task is to classify protein functions. 
PPI consists of 24 graphs, each corresponds to a human tissue. Each node has positional gene sets, motif gene sets, and immunological signatures as features and gene ontology sets as labels. 
20 graphs are used for training, 2 graphs are used for validation and the rest for test.

\noindent \textbf{DB Task.}
For the DB task, we choose the cross-lingual entity alignment, which matches entities referring to the same instances in different languages  in two KBs . In the literature~\cite{wang2018cross,xu2019cross}, GNN methods have been incorporated to this task to make use of the structure information underlying the cross-lingual KBs. We use the DBLP15K datasets built by \cite{sun2017cross}, which were generated from DBpedia, a large-scale multi-lingual KB containing rich inter-language links between different language versions. We choose the subset of Chinese and English in our experiments, and the statistics of the dataset are in Table~\ref{tb-dataset-db}.

For the experimental setting, we follow~\cite{wang2018cross} and use 30\% of inter-language links for training, 10\% for validation and the remaining 60\% for test. For the evaluation metric, we use $Hits@k$ to evaluate the performance of SANE. For the sake of space, we refer readers to~\cite{wang2018billion} for more details.

\subsubsection{Compared Methods}
We compare SANE with two groups of state-of-the-art methods: human-designed GNN architectures and NAS approaches.

\noindent
\textbf{Human-designed GNNs.}
As shown in Table~\ref{tb-gnn-comparison}, the human-designed GNN architectures are: GCN~\cite{kipf2016semi}, GraphSAGE~\cite{hamilton2017inductive}, GAT~\cite{velivckovic2017graph}, GIN~\cite{xu2018powerful}, LGCN~\cite{gao2018large}, and GeniePath~\cite{liu2018darts}. For models with variants, like different aggregators in GraphSAGE, we report the best performance across the variants. Besides, we add JK-Network to all models except for LGCN, and obtain 5 more baselines: GCN-JK, GraphSAGE-JK, GAT-JK, GIN-JK, GeniePath-JK. For LGCN, we use the code released by the authors\footnote{https://github.com/HongyangGao/LGCN}, and for other baselines, we use the popular open-source library PyTorch Geometric (PyG)~\cite{fey2019fast}, which implements various GNN models. 
For all baselines, we train it from scratch with the obtained best hyper-parameters on validation datasets, and get the test performance. We repeat this process 5 times, and report the final mean accuracy with standard deviation.

\begin{table}[t]
	\caption{The statistics of the dataset $\text{DBP15K}_{ZH-EN}$. }
	\setlength\tabcolsep{2.5pt}
	\centering
	\label{tb-dataset-db}
	\begin{tabular}{c|ccccc}
		\toprule
		& \#Entities & \#Relations & \#Attributes & \#Rel.triples & \#Attr.triples \\ \midrule
		Chinese & 66,469   & 2,830     & 8.113      & 153,929     & 379,684      \\ 
		English & 98,125   & 2,317     & 7,173      & 237,674     & 567,755      \\ \bottomrule
	\end{tabular}
\end{table}

\begin{table*}[ht]
	\centering
	\caption{Performance comparisons of transductive and inductive tasks. For transductive task, we use the mean classification accuracy (with standard deviation) as the evaluation metric, and for inductive task, we use Micro-F1 (with standard deviation) as evaluation metric. We categorize baselines into human-designed architectures and NAS approaches. The best results in different groups of baselines are underlined, and the best result on each dataset is in boldface.}
	\renewcommand{\arraystretch}{1.30}
	\begin{tabular}{c|l|ccc|c }
		\toprule
		&              & \multicolumn{3}{c|}{Transductive}          &  Inductive\\\cmidrule{2-6}
		&    Methods           & Cora            & CiteSeer        & PubMed          &  PPI\\\midrule
		\multirow{11}{65px}{Human-designed architectures}
		& GCN    & 0.8811 (0.0101) & 0.7666 (0.0202) & 0.8858 (0.0030) & 0.6500 (0.0000)  \\
		& GCN-JK        & 0.8820 (0.0118) & \underline{0.7763 (0.0136)} & 0.8927 (0.0037) & 0.8078(0.0000) \\
		& GraphSAGE          & 0.8741 (0.0159) & 0.7599 (0.0094) & 0.8834 (0.0044) & 0.6504 (0.0000) \\
		& GraphSAGE-JK       & \underline{0.8841 (0.0015)} & 0.7654 (0.0054) & \underline{0.8942 (0.0066)} & 0.8019 (0.0000) \\
		& GAT           & 0.8719 (0.0163) & 0.7518 (0.0145) & 0.8573 (0.0066) & 0.9414 (0.0000) \\
		& GAT-JK        & 0.8726 (0.0086) & 0.7527 (0.0128) & 0.8674 (0.0055) & \underline{0.9749 (0.0000)} \\
		& GIN           & 0.8600 (0.0083) & 0.7340 (0.0139) & 0.8799 (0.0046) & 0.8724 (0.0002) \\
		& GIN-JK        & 0.8699 (0.0103) & 0.7651 (0.0133) & 0.8878 (0.0054) & 0.9467 (0.0000) \\
		& GeniePath     & 0.8670 (0.0123) & 0.7594 (0.0137) & 0.8846 (0.0039) & 0.7138 (0.0000) \\
		& GeniePath-JK  & 0.8776 (0.0117)  & 0.7591 (0.0116) & 0.8868 (0.0037) & 0.9694 (0.0000)                \\
		& LGCN          & 0.8687 (0.0075) &       0.7543 (0.0221)           &     0.8753 (0.0012)            &  0.7720 (0.0020)               \\\midrule
		\multirow{5}{*}{NAS approaches}  & Random & 0.8594 (0.0072) & 0.7062 (0.0042) & 0.8866(0.0010)  &     0.9517 (0.0032)            \\
		& Bayesian      & 0.8835 (0.0072) & 0.7335 (0.0006) & 0.8801(0.0033)  &    0.9583 (0.0082)             \\
		& GraphNAS      & \underline{0.8840} (0.0071) & \underline{0.7762 (0.0061)} &       \underline{0.8896 (0.0024)}          &      \underline{0.9692 (0.0128)}           \\
		& GraphNAS-WS &     0.8808 (0.0101)            &    0.7613 (0.0156)             & 0.8842 (0.0103)                &    0.9584 (0.0415)             \\\midrule
		one-shot NAS& SANE          & \textbf{0.8926 (0.0123)}      &        \textbf{0.7859 (0.0108)}         & \textbf{0.9047 (0.0091)}               & \textbf{0.9856 (0.0120)} \\\bottomrule
	\end{tabular}
	\label{tb-performance-trans-induc}
\end{table*}

\noindent
\textbf{NAS approaches for GNN.}
We consider the following methods:
(i).
Random search (denoted as ``Random'') \cite{bergstra2012random}: 
a simple baseline in NAS, which uniform randomly samples architectures from the search space;
(ii). 
Bayesian optimization\footnote{\url{https://github.com/hyperopt/hyperopt}}
(denoted as ``Bayesian'') \cite{bergstra2011algorithms}:
a popular sequential model-based global optimization method
for hyper-parameter optimization,
which uses tree-structured Parzen estimator as the measurement for expected improvement;
(iii).
GraphNAS\footnote{\url{https://github.com/GraphNAS/GraphNAS}}~\cite{gao2019graphnas}, 
a RL-based NAS approach for GNN, which has two variants based on the adoption of weight sharing mechanism. 
We denoted as GraphNAS-WS the one using weight sharing. Note that Auto-GNN~\cite{zhou2019auto} is not compared for three reasons: 1) the search spaces of Auto-GNN and GraphNAS are actually the same; 2) both of these two works use the RL method; 3) the code of Auto-GNN is not publicly available. 

Random and Bayesian are searching on the designed search space of SANE, where a GNN architecture is sampled from the search space, and trained till convergence to obtain the validation performance. 200 models are sampled in total and the architecture with the best validation performance is trained from scratch, and do some hyper-parameters tuning on the validation dataset, and obtain the test performance. For GraphNAS, we set the epoch of training the RL-based controller to 200, and in each epoch, a GNN architecture is sampled, and trained for enough epochs ($600\sim1000$ depending on datasets), update the parameters of RL-based controller. In the end, we sample 10 architectures and collect the top 5 architectures that achieve the best validation accuracy. Then the best architecture is trained from scratch. Again, we do some hyper-parameters tuning based on the validation dataset, and report the best test performance. Note that we repeat the re-training of the architecture for five times, and report the final mean accuracy with standard deviation.

\subsubsection{Implementation details of SANE}
Our experiments are running with Pytorch (version 1.2)~\cite{paszke2019pytorch} on a GPU 2080Ti (Memory: 12GB, Cuda version: 10.2). We implement SANE on top of the building code provided by DARTS\footnote{https://github.com/quark0/darts} and PyG (version 1.2)\footnote{https://github.com/rusty1s/pytorch\_geometric}. More implementing details are given in Appendix~\ref{append-hyper-paras-gnn}. Note that we set $\xi = 0$ in \eqref{eq:grad} in our experiments, which means we are using first-order approximation as introduced in~\cite{liu2018darts}. It is more efficient and the performance is good enough in our experiments.
For all tasks, we run the search process for 5 times with different random seeds, and retrieve top-1 architecture each time. By collecting the best architecture out of the 5 top-1 architectures on validation datasets, we repeat 5 times the process in re-training the best one, 
fine-tuning hyper-parameters on validation data, and reporting the test performance. Again, the final mean accuracy with standard deviations are reported.

\begin{figure*}[ht]
	\centering
	\subfigure[Cora.]{\includegraphics[width=0.21\textwidth]{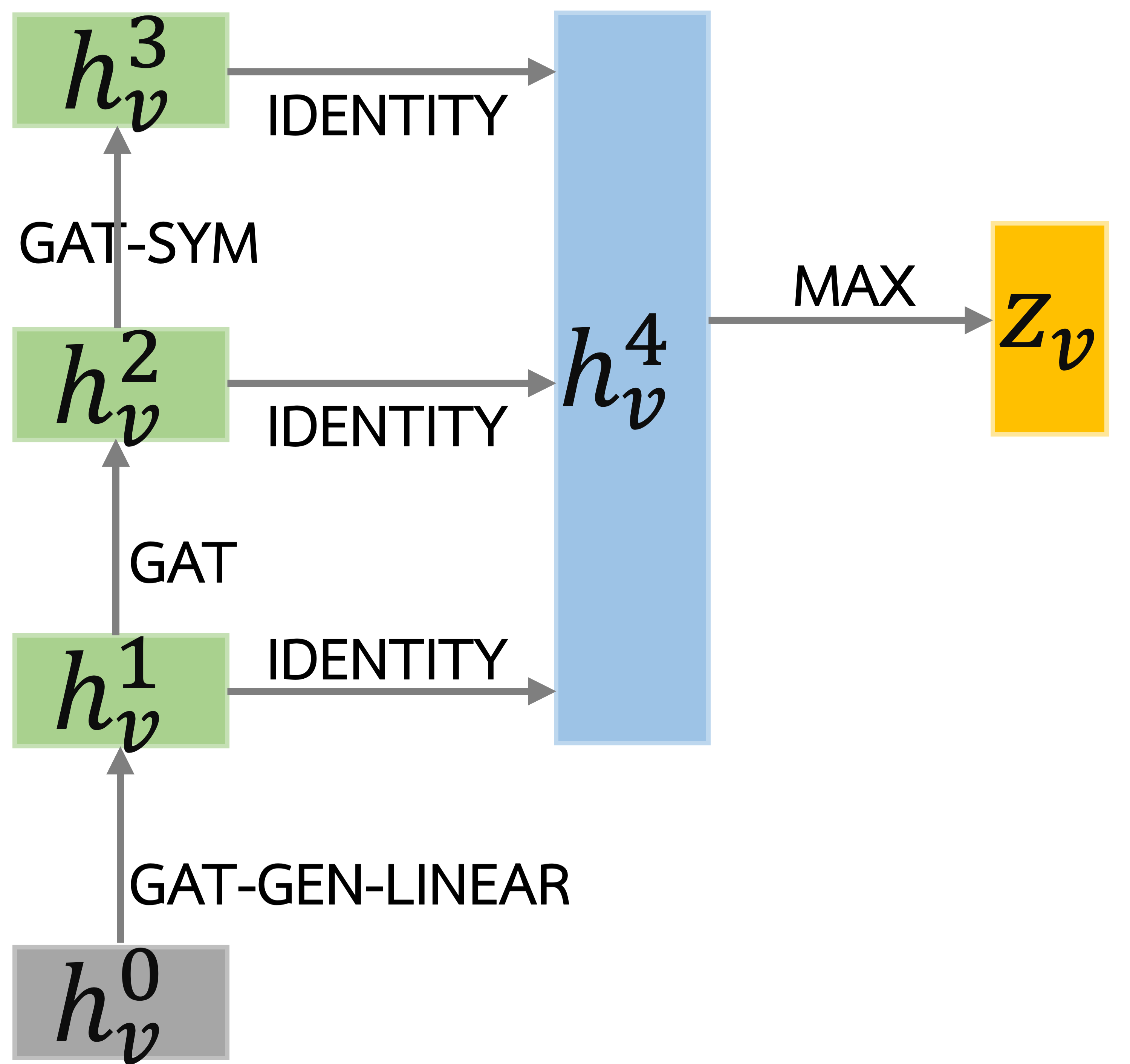}}
	\subfigure[CiteSeer.]{\includegraphics[width=0.21\textwidth]{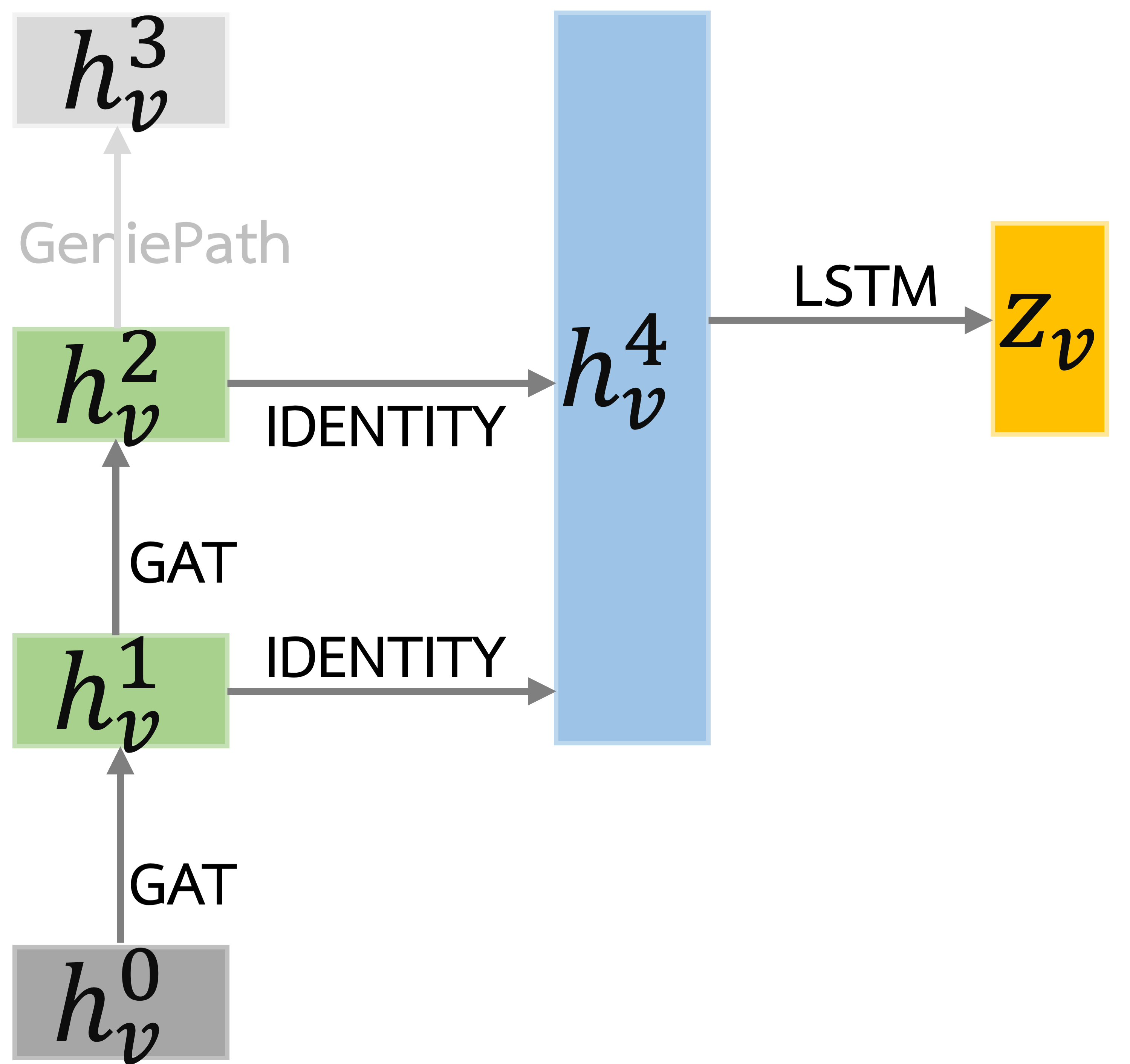}}
	\subfigure[PubMed.]{\includegraphics[width=0.21\textwidth]{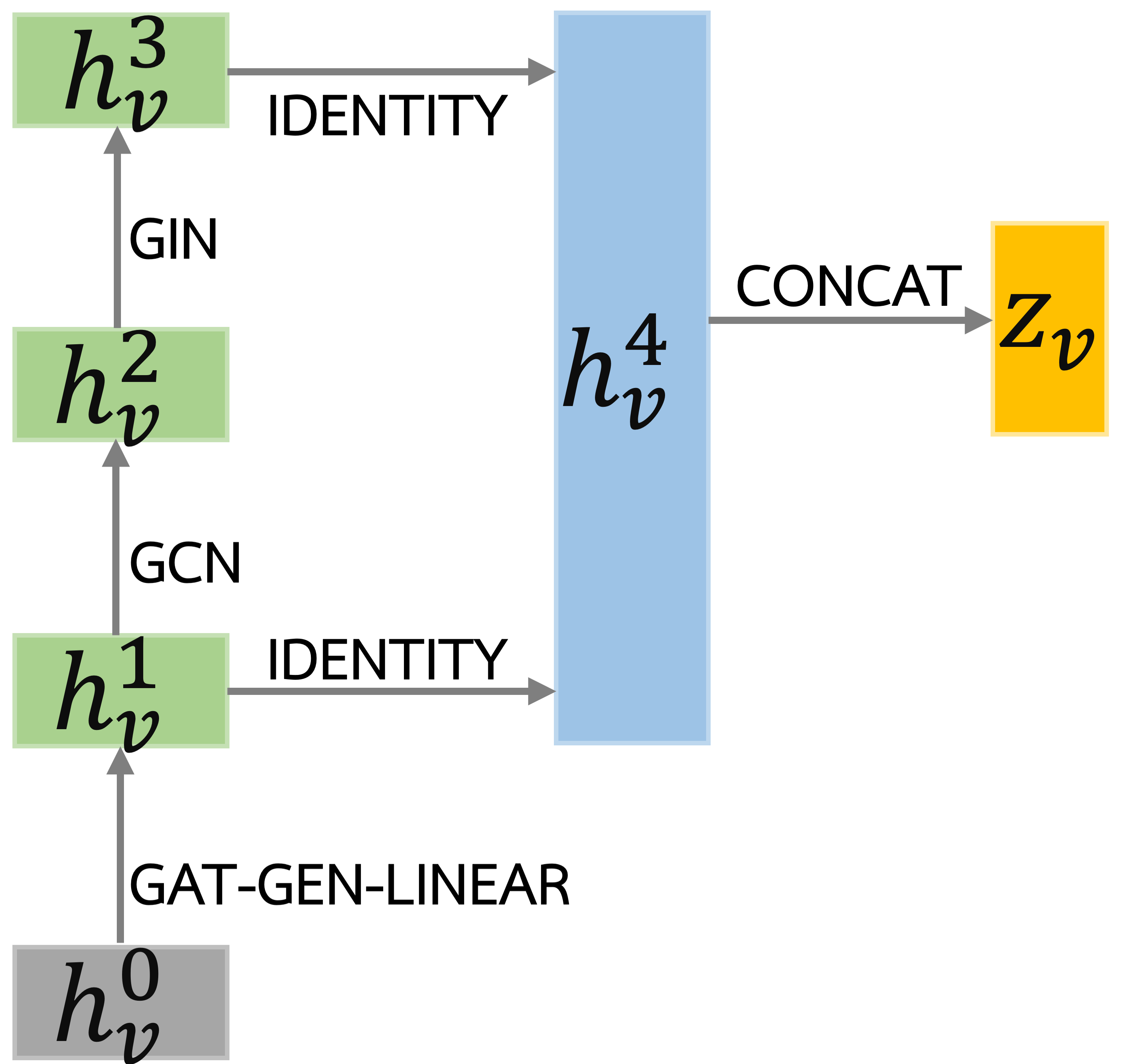}}
	\subfigure[PPI.]{\includegraphics[width=0.21\textwidth]{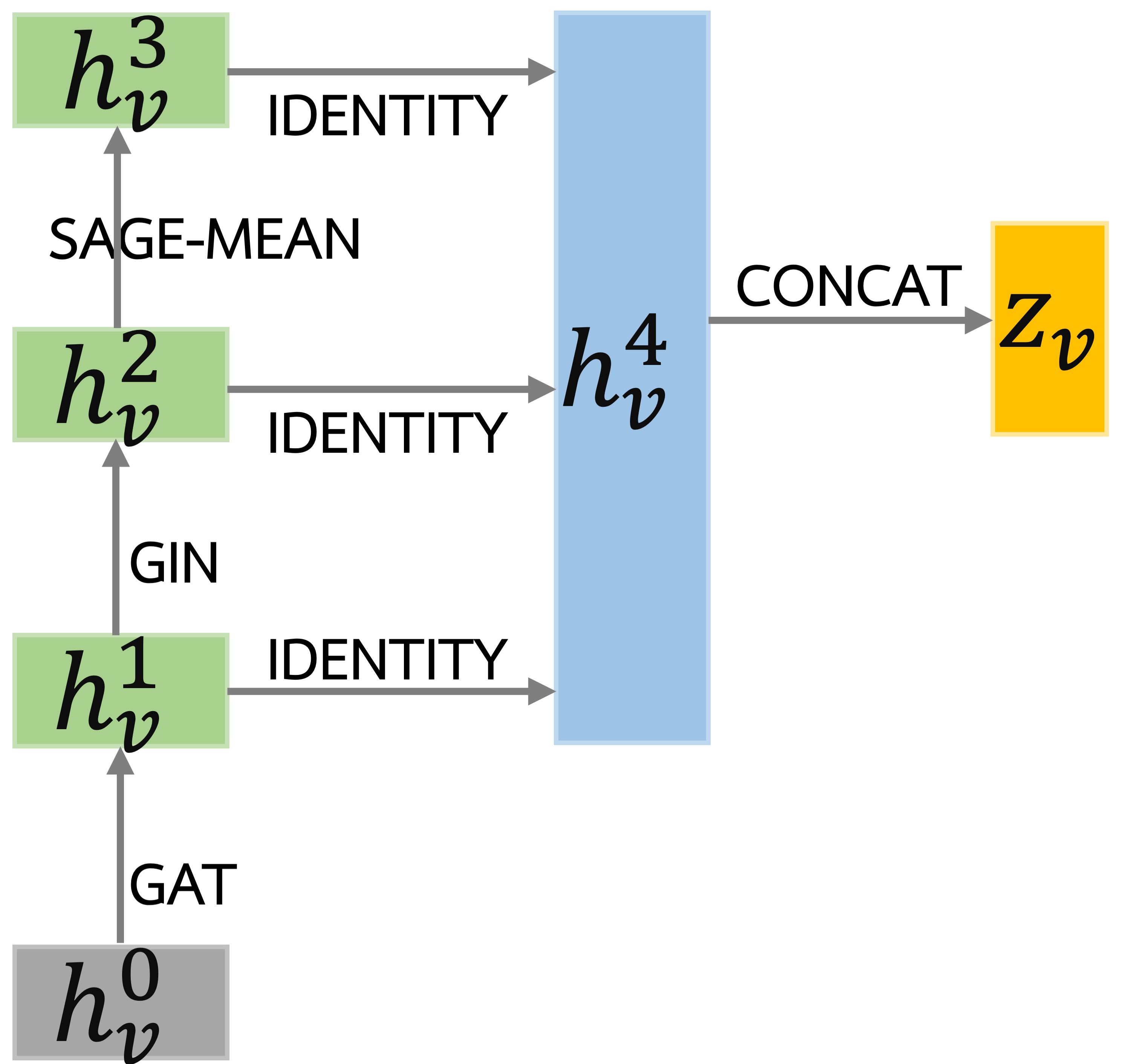}}
	\caption{The searched architectures by SANE on different datasets. Note that on CiteSeer, 
		the skip connection is removed during the search process, 
		thus we make it in light gray for clear presentation.}
	\label{fig-case-trans-induc}
\end{figure*}

\subsection{Performance on Transductive and Inductive Tasks}
\label{sec-exp-performance}

The results of transductive and inductive tasks are given in Table~\ref{tb-performance-trans-induc}, 
and we give detailed analyses in the following.

\subsubsection{Transductive Task} 
Overall, we can see that SANE consistently outperforms all baselines on three datasets, 
which demonstrates the effectiveness of the searched architectures by SANE.
When looking at the results of human-designed architectures, 
we can first observe that
 GCN and GraphSAGE outperform other more complex models, e.g., GAT or GeniePath, which is similar to the observation in the paper of GeniePath~\cite{liu2019geniepath}.
We attribute this to the fact that these three graphs are not that large, 
thus the complex aggregators might be easily prone to the overfitting problem.
Besides, there is no absolute winner among human-designed architectures, which further verifies the need of searching for data-specific \mbox{architectures}.
Another interesting observation is that when adding JK-Network to base GNN architectures, the performance increases consistently, which aligns with the experimental results in JK-Network~\cite{xu2018representation}. 
It demonstrates the importance of introducing the layer aggregators into the search space of SANE.

On the other hand, when looking at the performance of NAS approaches, 
the superiority of SANE is also clear from the gain on performance. 
Recall that, Random Bayesian and GraphNAS all search in a discrete search space,
while SANE searches in a differentiable one enabled by Equation~\eqref{eq-continuous-relaxation}.
This shows a differentiable search space is easier for search algorithms to find a better local optimal.
Such an observation is also previously made in \cite{liu2018darts,yao2019differentiable}, 
which search for CNN in a differentiable space.

\subsubsection{Inductive Task} 
We can see a similar trend in the inductive task that SANE performs consistently better than all baselines. 
However, 
among human-designed architectures, 
the best two are GAT and GeniePath with JK-network,
which is not the same as that from the transductive task.
This further shows the importance to search data-specific GNN architectures.


\subsubsection{Searched Architectures} 
We visualize the searched architectures (top-1) by SANE on different datasets in Figure~\ref{fig-case-trans-induc}. 
As can be seen,
first, these architecture are data-dependent and new to the literature.
Then, searching for skip-connections indeed make a difference,
as the last GNN layer in Figure~\ref{fig-case-trans-induc}(b) and
the middle layer in Figure~\ref{fig-case-trans-induc}(c) do not connect to the output layer.
Finally, as attention-based node aggregators are more expressive than non-attentive ones,
thus \texttt{GAT} (and its variants) are more popularly used.


\begin{figure*}[ht]
	\subfigure[Cora.]{\includegraphics[width=0.24\textwidth]{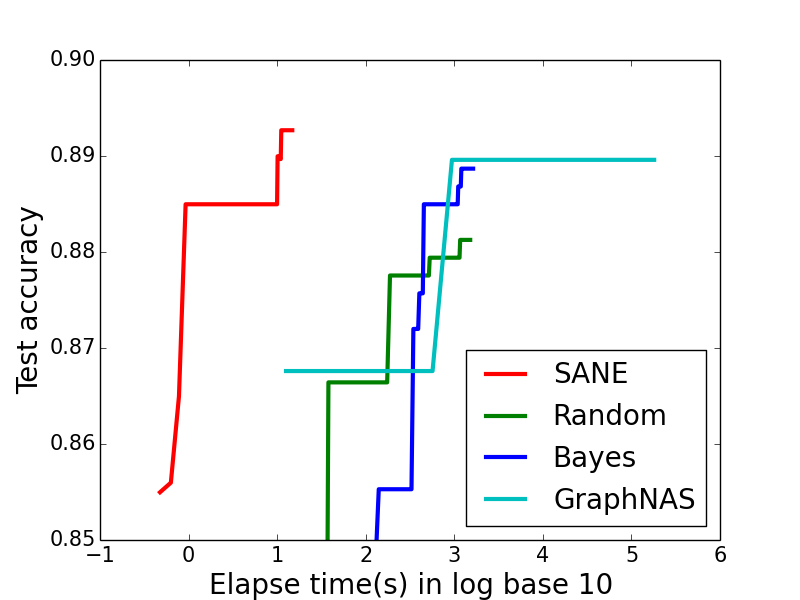}}
	\subfigure[CiteSeer.]{\includegraphics[width=0.24\textwidth]{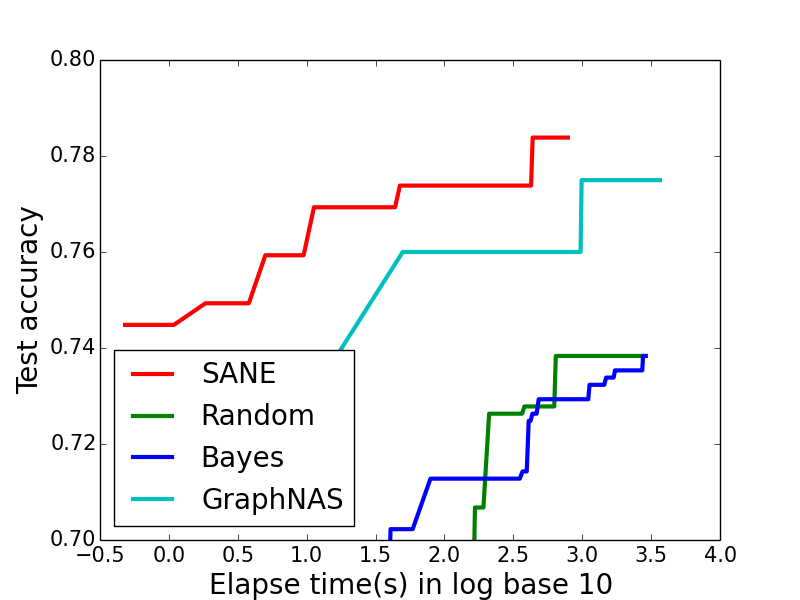}}
	\subfigure[PubMed.]{\includegraphics[width=0.24\textwidth]{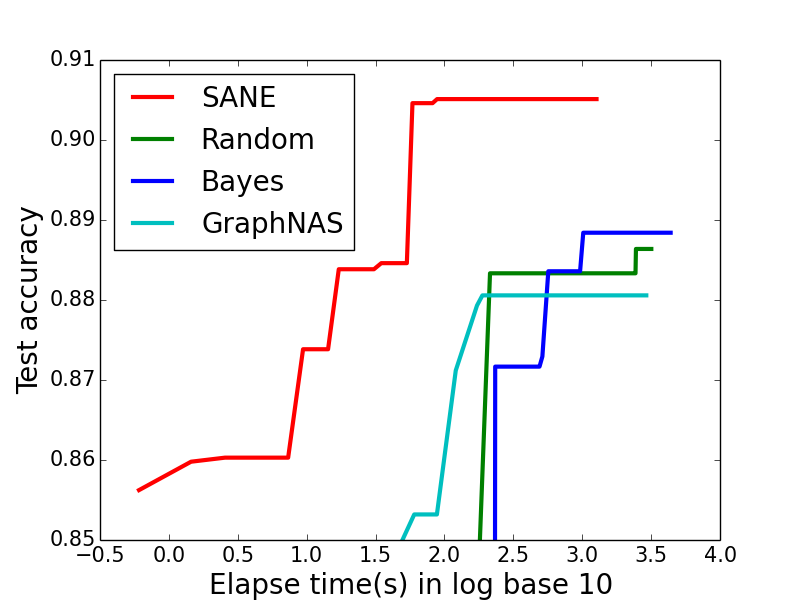}}
	\subfigure[PPI.]{\includegraphics[width=0.24\textwidth]{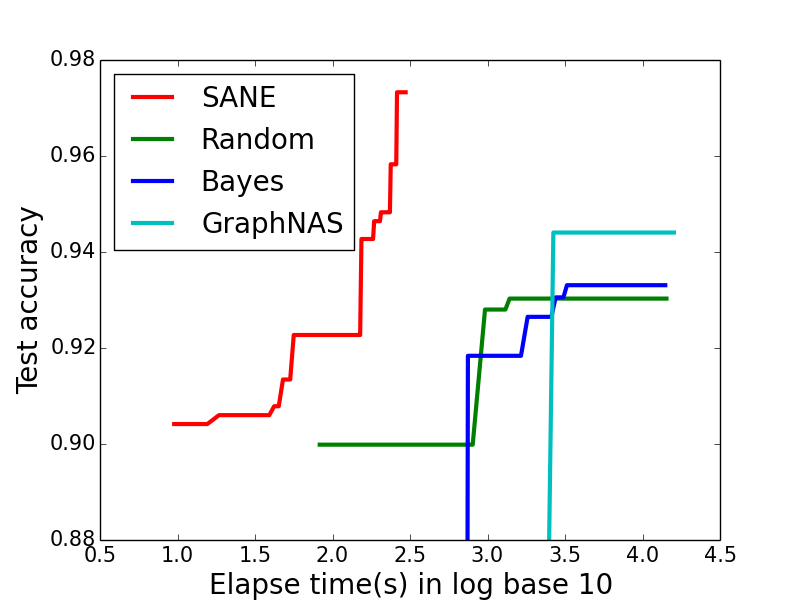}}
	\centering
	\caption{The test accuracy w.r.t. search time (in seconds) in log base 10.}
	\label{fig-search efficiency}
\end{figure*}

\subsection{Search Efficiency}
\label{sec-exp-search-efficiency}

In this part, we compare the efficiency of SANE and NAS baselines by showing the test accuracy w.r.t the running time on transductive and inductive tasks.
From Figure~\ref{fig-search efficiency},  
we can observe that the efficiency improvements are in orders of magnitude, which aligns with the experiments in previous one-shot NAS approaches, like DARTS~\cite{liu2018darts} and NASP~\cite{yao2019differentiable}.
Besides,
as in Section~\ref{sec-exp-performance},
we can see that SANE can obtain an architecture with higher test accuracy than random search, Bayesian, and GraphNAS.

To further show the efficiency improvements, 
we record the search time of each method to obtain an architecture, 
where the epochs of SANE and GraphNAS are set to 200, and the number of trial-and-error process in Random and Bayesian is set to 200 as well, i.e., explore 200 candidate architectures, and the results are given in Table~\ref{tb-total-cost}. 
The search time cost of SANE is two orders of magnitude smaller than those of NAS baselines, which further demonstrates the superiority of SNAE in terms of search efficiency.

\begin{table}[ht]
		\centering
	\caption{The search time (clock time in seconds) of running once each mode.}
	\label{tb-total-cost}
	\begin{tabular}{l|ccc|c}
		\toprule
		& \multicolumn{3}{c|}{Transductive task} & \multicolumn{1}{c}{Inductive task} \\\cmidrule{2-5}
		& Cora      & CiteSeer     & PubMed     & PPI                                \\\midrule
		Random & 1,500      & 2,694         & 3,174       & 13,934                              \\
		Bayesian      & 1,631      & 2,895         & 4,384       & 14,543                              \\
		GraphNAS      & 3,240      & 3,665         & 5,917       & 15,940                              \\
		SANE          & 14        & 35           & 54         & 298                               \\\bottomrule
	\end{tabular}
\end{table}

\begin{table}[ht]
	\centering
	\caption{The results of DB task. We use Hits@K as the evaluation metric, and the results of JAPE and GCN-Align are from~\cite{wang2018cross}. Note that JAPE is the variant using the same features as GCN-Align.}
	\label{tb-exp-db-task}
	\begin{tabular}{c|ccc|ccc}
		\toprule
		& \multicolumn{3}{c|}{ZH$\rightarrow$EN}  & \multicolumn{3}{c}{EN$\rightarrow$ZH}  \\ \cmidrule{2-7}
		& @1 & @10 & @50 & @1 & @10 & @50 \\ \midrule
		JAPE & 33.32  & 69.28   & 86.40   & 33.02  & 66.92   & 85.15   \\
		GCN-Align  & 41.25  & 74.38   & 86.23   & 36.49  & 69.94   & 82.45   \\\midrule
		SANE & \textbf{42.10}  & \textbf{74.51}  &  \textbf{88.12}       &  \textbf{38.41}     &  \textbf{70.23}      &  \textbf{85.43}       \\ \bottomrule
	\end{tabular}
\end{table}

\begin{table*}[ht]
	\centering
	\caption{Performance comparisons of two search spaces on four benchmark datasets. We show the mean classification accuracy (with STD). ``-WS'' represents the weight sharing variant of GraphNAS.}
	\begin{tabular}{L{128px}|cccc}
		\toprule
		Methods                        &           Cora           &         CiteSeer         &          PubMed          &           PPI            \\ \midrule
		GraphNAS                       &     0.8840 (0.0071)      & \textbf{0.7762 (0.0061)} &     0.8896 (0.0024)      &     0.9698 (0.0128)      \\
		GraphNAS-WS                    &     0.8808 (0.0101)      &     0.7613 (0.0156)      &     0.8842 (0.0103)      &     0.9584 (0.0415)      \\ \midrule
		GraphNAS(SANE search space)    &     0.8826 (0.0023)      &     0.7707 (0.0064)      &     0.8877 (0.0012)      & \textbf{0.9887 (0.0010)} \\
		GraphNAS-WS(SANE search space) & \textbf{0.8895 (0.0051)} &     0.7695 (0.0069)      & \textbf{0.8942 (0.0010)} &     0.9875 (0.0006)      \\ \bottomrule
	\end{tabular}
	\label{tb-exp-snag}
\end{table*}

\subsection{DB task} 
\label{sec:exp:db}
Since DB task is different from the benchmark tasks, we adjust the settings of SANE following the proposed GCN-Align~\cite{wang2018cross}, 
which uses two 2-layer GCN in their experiments. To be specific, we set the number of layers to 2, which is different from that in the transductive task, and remove the layer aggregator because we observe that in our experiments the performance decreases when simply adding the layer aggregator to the GCN architecture in~\cite{wang2018cross}. Therefore, we use SANE to search for different combinations of node aggregators for the entity alignment task, and the results are shown in Table~\ref{tb-exp-db-task}. We can see that the performance of SANE is better than GCN-Align and  JAPE, which demonstrates the effectiveness of SANE on the entity alignment task. We further emphasize the following observations:
\begin{itemize}[leftmargin=*]
\item The performance gains of SANE compared to GCN-Align is evident, 
which demonstrates the advantage of different combinations of node aggregators. 
And The searched architecture is ``\texttt{GAT}-\texttt{GeniePath}'', and more hyper-parameters are given in Appendix~\ref{append-hyper-paras-gnn}.	


\item The experimental results show the capability of SANE in broader domains. 
A taking-away message is that SANE can further improve the performance of a task where a regular GNN model, e.g., GCN, can work.
\end{itemize}
We notice that there are following works of GCN-Align, e.g., \cite{xu2019cross,cao2019multi}, however their modifications are orthogonal to node aggregators, thus they can be integrated with SANE to further improve the performance. We leave this for future work.

\subsection{Ablation Study}

\subsubsection{The influence of differentiable search} 
\begin{figure}[ht]
	\centering
	\subfigure[Random explore: $\epsilon$.]{\includegraphics[width=0.49\columnwidth]{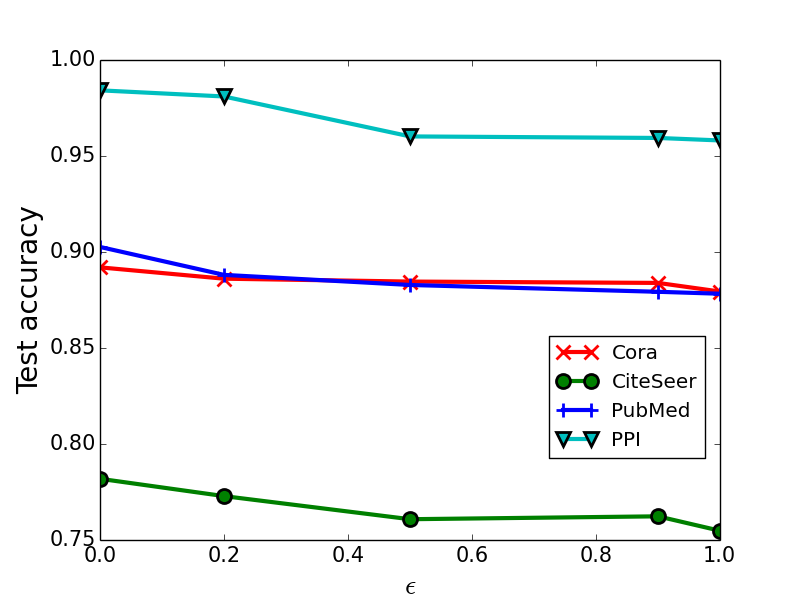}}
	\subfigure[Number of GNN layers: $K$.]{\includegraphics[width=0.49\columnwidth]{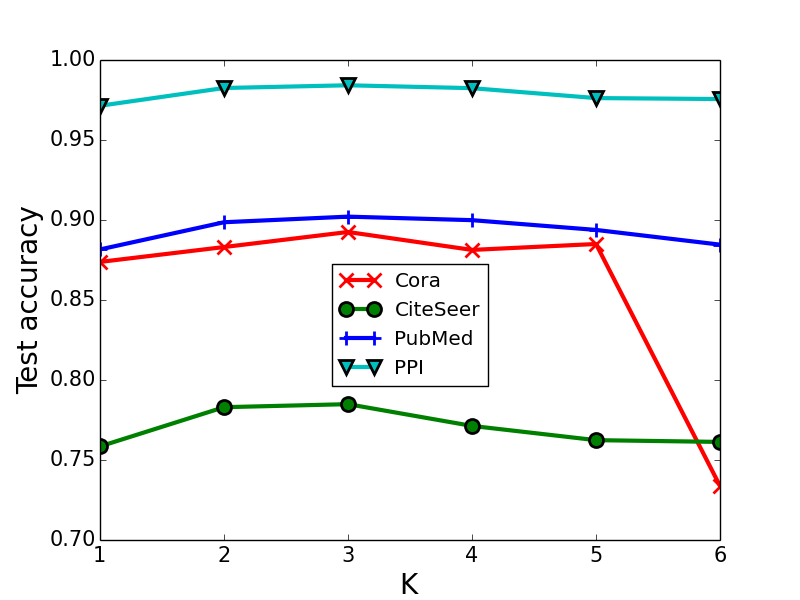}}
	\caption{The test accuracy w.r.t. different $\epsilon$ and $K$.}
	\label{fig-ablation}
\end{figure}

In Algorithm~\ref{alg-sane} in Section~\ref{sec-framework-search-algorithm}, we can see that during the search process, the $\bm{\alpha}$ is updated w.r.t. to the validation loss $\mathcal{L}$ and used to update the weights of the mixed operations in \eqref{eq-sane-con-formula-1} to \eqref{eq-sane-con-formula-2}. Here we introduce a random explore parameter $\epsilon$, which is the probability that we randomly sample a single operation in each edge of the supernet, and only update the weights corresponding to the sampled operations. Then when $\epsilon = 0$, the algorithm is the same to Algorithm~\ref{alg-sane}, 
and when $\epsilon = 1$, 
it is equivalent to random search with weight sharing.
Thus we can show the influence of the differentiable search algorithm by varying $\epsilon$. 
In this part, we conduct experiments by varying $\epsilon \in [0, 0.2, 0.5, 0.9, 1.0]$, and show the performance trending in Figure~\ref{fig-ablation}(a) on the transductive and inductive tasks.

From Figure~\ref{fig-ablation}(a), on all four datasets, we can see that the test accuracy decreases with the increasing of $\epsilon$, and arrives the worst when $\epsilon = 1.0$. This means that the gradient descent method outperforms random sampling for architecture search. In other words, it demonstrates the effectiveness of the proposed differentiable search algorithm in Section~\ref{sec-framework-search-algorithm}.

\subsubsection{The influence of $K$} 
In Section~\ref{sec-exp}, we choose a $3$-layer GNN (K=3) as the backbone in our experiments for its empirically good performance. In this work, we focus on searching for shallow GNN architectures ($K \leq 6$). In this part, we conduct experiments by varying $K \in [1,2,3,4,5,6]$, and show the performance trending in Figure~\ref{fig-ablation}(b).
We can see that with the increase of $K$, the test accuracy increases firstly and then decreases, which aligns with the motivation of setting $K=3$ in the experiments in Section~\ref{sec-exp-performance}.

\subsubsection{The efficacy of the designed search space}
\label{sec-exp-abl-search-space}
In Section \ref{sec-framework-graphnas-comparison},  we discuss the advantages of search space between SANE and GraphNAS/Auto-GNN. In this part, we conduct experiments to further show the advantages. To be specific, we run GraphNAS over its own and SANE's search space, given the same time budget (20 hours), and compare the final test accuracy of the searched architectures in Table~\ref{tb-exp-snag}. From Table~\ref{tb-exp-snag}, we can see that despite the simplicity of the search space, SANE can obtain better or at least close accuracy compared to GraphNAS, which means better architectures can be obtained given the same time budget, thus demonstrating the efficacy of the designed search space.

\subsubsection{Failure of searching for universal approximators} 
\label{sec:MLP}
In this part, we further show the performance of searching Multi-Layer Perception (MLP) as node aggregators since it is a universal function approximator: any continuous function on a compact set can be approximated with a large enough MLP~\cite{leshno1993multilayer}. 
And in~\cite{xu2018powerful}, Xu et al. shows that using MLP aggregators can be as powerful as the Weisfeiler-Lehman (WL) test under some conditions, which upper bounds the expressive capability of existing GNN models. However, in practice, it is very challenging to obtain satisfying performance by MLPs without prior knowledge, since it is too general to design the MLP of a suitable structure given a task. This motivates us to incorporate various existing GNN models as node aggregators in our search space (Table~\ref{tb-search-space-ops}), thus the performance of any search algorithm can be guaranteed. Then to show the challenges of directly searching for MLP as node aggregators, we propose to search for a specific MLP as node aggregators with NAS approaches. To be specific, we set the parameter space of the MLP as $w \in [8, 16, 32, 64]$ and $d \in [1, 2 ,3]$, where $w$ and $d$ represents the hidden embedding size (width) and the depth of the MLP, respectively. For simplicity, we adopt the Random and Bayesian as the NAS approaches, and search on the four benchmark datasets, where the settings are the same as those in Section~\ref{sec-exp-performance}. 
The performance is shown in Table~\ref{tb-mlp-performance}, from which we can see that the performance gap between searching for MLP and SANE are large. 
This observation demonstrates the difficulties of searching MLP as node aggregators despite its powerful expressive capability as WL test. On the other hand, it demonstrates the necessity of the designed search space of SANE, which includes existing human-designed GNN models, thus the performance can be guaranteed in practice.

\begin{table}[t]
	\caption{The performance of searching for MLP. We list the best performance of SANE from Table~\ref{tb-performance-trans-induc} as comparison.}
	\centering
	\label{tb-mlp-performance}
	\begin{tabular}{c|cc|c}
		\toprule
		Dataset         & Random  & Bayesian  & SANE  \\ \midrule
		Cora           & 0.8698 (0.0011) &  0.8470 (0.0032) & \textbf{0.8926 (0.0123)}        \\
		CiteSeer        &  0.7298 (0.0078) & 0.7103 (0.0057)&      \textbf{0.7859 (0.0108)}  \\
		PubMed           & 0.8662 (0.0030)&  0.8699 (0.0065) &   \textbf{0.9047 (0.0091)}    \\\midrule
		PPI           &0.8166 (0.0089)  & 0.8685 (0.0017)  &    \textbf{0.9856 (0.0120)}    \\
	 \bottomrule
	\end{tabular}
\end{table}

\section{Conclusion}
\label{sec-conclusion}
In this work, to address the architecture and computational challenges facing existing GNN models and NAS approaches, we propose to Search to Aggregate NEighborhood (SANE) for graph neural architecture search. By reviewing various human-designed GNN architectures, we define an expressive search space including node and layer aggregators, which can emulate more unseen GNN architectures beyond existing ones. A differentiable architecture search algorithm is further proposed, which leads to a more efficient search algorithm than existing NAS methods for GNNs.
Extensive experiments are conducted on five real-world datasets in transductive, inductive, and DB tasks. The experimental results demonstrate the superiority of SANE comparing to GNN and NAS baselines in terms of effectiveness and efficiency.

For future directions, we will explore more advanced NAS approaches to further improve the performance of SANE. 
Besides, we can explore beyond node classification tasks and focus on more graph-based tasks, 
e.g., the whole graph classification \cite{hu2020open}.
In these cases, 
different graph pooling methods can be searched for the whole graph representations.

\section{Acknowledgments}
This work is supported by National Key R\&D Program of China (2019YFB1705100). We also thank Lanning Wei to implement several experiments in this work. We further thank all anonymous reviewers for their constructive comments, which help us to improve the quality of this manuscript.

\begin{table*}[ht]
	\centering
	\caption{More explanations to the node aggregators in \eqref{eq-mpnn}.}
	\label{tb-node-agg-detail}
	\begin{tabular}{l|C{100px}|l}
		\toprule
		GNN models                                &           Symbol in the paper            & Key explanations                                                                                      \\ \midrule
		GCN~\cite{kipf2016semi}                 &                   \texttt{GCN}                    & $F^l_N (v) = \sum\limits_{u \in \widetilde{N}(v)}\big(\text{degree}(v)\cdot \text{degree}(u)\big)^{-1/2}\cdot \bh_u^{l-1} $ \\ \midrule
		GraphSAGE~\cite{hamilton2017inductive}           & \texttt{SAGE-MEAN}, \texttt{SAGE-MAX}, \texttt{SAGE-SUM} & Apply mean, max, or sum operation to $\{\bh_u |u \in \widetilde{N}(v)\}$.                      \\ \midrule
		\multirow{5}{*}{GAT~\cite{velivckovic2017graph}} 
		&                   \texttt{GAT}      & Compute attention score: $\be^{gat}_{uv} =  \text{Leaky\_ReLU}\big(\ba\big[\bW_u\bh_u||\bW_v\bh_v\big]\big)$. \\ \cmidrule{2-3}
		&                 \texttt{GAT-SYM}                  &  $\be^{sys}_{uv} = \be^{gat}_{uv} + \be^{gat}_{vu}  $.      \\ \cmidrule{2-3}
		&                 \texttt{GAT-COS}                  &    $\be^{cos}_{uv} = \langle\bW_u\bh_u, \bW_v\bh_v\rangle  $.        \\ \cmidrule{2-3}
		&               \texttt{GAT-LINEAR}              &   $\be^{lin}_{uv} = \text{tanh}\big(\bW_u\bh_u + \bW_v\bh_v\big) $. \\ \cmidrule{2-3}
		&              \texttt{GAT-GEN-LINEAR}  & $\be^{gen-lin}_{uv} = \bW_G\text{tanh}\big(\bW_u\bh_u + \bW_v\bh_v\big) $.  \\ \midrule
		GIN~\cite{xu2018powerful} &  \texttt{GIN}   &  $F^l_N (v) = \text{MLP}\bigg((1+\epsilon^{l-1}) \cdot \bh_v^{l-1} + \sum\limits_{u \in N(v)}\bh_u^{l-1} \bigg)$.                                                                                                  \\ \midrule
		LGCN~\cite{gao2018large}  &\texttt{CNN} &         Use 1-D CNN as the aggregator, equivalent to a weighted summation aggregator.                                                                                        \\ \midrule
		GeniePath~\cite{liu2019geniepath}&   \texttt{GeniePath}   &  Composition of GAT and LSTM-based aggregators \\ \midrule
		JK-Network~\cite{xu2018representation} &           &    Depending on the base above GNN\\ \bottomrule
	\end{tabular}
\end{table*}

\begin{table*}[ht]
	\centering
	\caption{The hyper-parameters obtained by hyperopt in the fine-tuning process for the searched architectures.}
	\label{tb-sane-hyper-paras}
	\begin{tabular}{l|c|c|c|c|c}
		\toprule
		&Cora&CiteSeer& PubMed & PPI & $\text{DBP15K}_{ZH-EN}$\\ \midrule
		Head num&8  & 2   &  2 & 4  &4\\\midrule	 
		Hidden embedding size&256  & 64   & 64  & 512  &512\\\midrule	 
		Learning rate&4.150e-4  & 5.937e-3   & 2.408e-3  & 1.002e-3  &2.039e-3\\\midrule	 
		$L_2$ Norm&1.125e-4  &  2.007e-5  & 8.850e-5  &  0 & 3.215e-4\\ \midrule
		Activation function &relu  & relu   & relu  & relu  &relu\\
		\bottomrule
	\end{tabular}
\vspace{-12pt}
\end{table*}

\bibliographystyle{ieeetr}
\bibliography{main}

\appendix

\subsection{Discussion about recent GNN baselines}
\label{append-baseline}
During the period of preparing for this work, we noticed that there were some new GNN models proposed in the literature, e.g., Geom-GCN~\cite{pei2020geom}, GraphSaint~\cite{zeng2019graphsaint}, DropEdge~\cite{rong2019dropedge}, and PariNorm~\cite{zhao2020pairnorm}. We did not include these works as baselines, since they can be regarded as orthogonal works of SANE to the GNN literature. 

To be specific, as shown in Eq.~\eqref{eq-mpnn}, the embedding of a node $v$ in the $l$-th layer of a $K$-layer GNN is computed as:
\begin{align*}
\bh_v^l =  \sigma(\bW^{l} \cdot \text{AGG}_{\text{node}}(\{\bh_u^{l-1}, \forall u \in \widetilde{N}(v)\})).
\end{align*}
From this computation process, we can summarize four key components of a GNN model: \textit{aggregation function ($\text{AGG}_{\text{node}}$)}, \textit{number of layers ($l$)}, \textit{neighbors ($\widetilde{N}(v)$)}, and \textit{hyper-parameters ($\sigma$, dimension size, etc.)}, which decide the properties of a GNN model, e.g., the model capacity, expressive capability, and prediction performance. 

SANE mainly focus on the aggregation functions, which affect the expressive capability of GNN models. GraphSaint mainly focuses on neighbors selection in each layer, thus the ``neighbor explosion'' problem can be addressed. Geom-GCN also focuses on neighbors selection, which constructs a novel neighborhood set in the continuous space, thus the structural information can be utilized. DropEdge mainly focuses on the depth of a GNN model, i.e., the number of layers, which can alleviate the over-smoothing problem with the increasing of the number of GNN layers. Besides the three works, there are more other works on the four key components, like MixHop~\cite{abu2019mixhop} integrating neighbors of different hops in a GNN layer, or PairNorm~\cite{zhao2020pairnorm} working on the depth of a GNN models. Therefore, all these works can be integrated as a whole to improve each other. For example, the DropEdge or Geom-GCN methods can further help SANE in constructing more powerful GNN models. This is what we mean ``orthogonal'' works of SANE. Since we mainly focus on the aggregation functions in this work, we only compare the GNN variants with different aggregations functions.
We believe the application of NAS to GNN has unique values, and the proposed SANE can benefit the GNN community.

\subsection{Details of Node Aggregators}
\label{append-node-agg}
As introduced in Section~\ref{sec-framework-search-space}, we have 11 types of node aggregators, which are based on well-known existing GNN models: GCN~\cite{kipf2016semi}, GraphSAGE~\cite{hamilton2017inductive}, GAT~\cite{velivckovic2017graph}, GIN~\cite{xu2018powerful}, and GeniePath~\cite{liu2019geniepath}. 
Here we give key explanations to these node aggregators in Table~\ref{tb-node-agg-detail}. For more details, we refer readers to the original papers.

\begin{table}[ht]
	\centering
	\caption{More implementing details of GNN baselines. Here we give \textit{the number of layers, hidden dimension size, activation function}, and \textit{the number of heads (GAT models)}. For JK-network, we further give the layer aggregators.}
	\label{tb-gnn-impl-detail}
	\begin{tabular}{l|c|c|c}
		\toprule
		&Cora\&CiteSeer                          & PubMed & PPI \\ \midrule
		GCN  &  $3, 64$, elu   & $3, 128$, elu  &  $3, 256$, elu\\ \midrule
		GraphSAGE &  $2, 64$, relu& $2, 128$, relu       & $3, 256$, elu    \\ \midrule
		GAT &$3, 64$, relu, 8  &   $3, 128$, relu, 8  &  $3, 256$, relu, 8   \\ \midrule
		GIN            & $3, 128$, relu &  $3, 128$, relu       &   $3, 256$, relu   \\ \midrule
		LGCN             & $3, 128$, relu &    $3, 128$, relu     & $3, 256$, relu     \\ \midrule
		GeniePath & $3, 256$, tanh, 4 &$3, 256$, tanh, 4  & $3, 256$, tanh, 4 \\ \midrule
		JK-Network &  CONCAT &    CONCAT    &  LSTM   \\ \bottomrule
	\end{tabular}
\vspace{-10pt}
\end{table}

\subsection{More Implementing Details}
\label{append-hyper-paras-gnn}
In this part, we give more implementation details of all methods including GNN baselines, NAS baselines, and SANE. 


\begin{itemize}[leftmargin=*]
	
	\item For all GNN baselines, we use the $Adam$ optimizer, and set learning rate $lr=0.005$, dropout $p=0.5$, and $L_2$ norm to $0.0005$. For other parameters, we do some tuning, and present the best ones in Table~\ref{tb-gnn-impl-detail}.
	
	\item For Random and Bayesian, the number of searched architectures is set to 200, and for each sampled architecture, we tune it using hyperopt for 50 iterations.
	
	\item For SANE, in the search phase, hidden embedding size is set to $32$ for sake of computational resource, and $lr=0.005$, dropout $p=0.6$, and $L_2$ norm $= 0.0002$, and for each searched architecture, we tune it using hyperopt\footnote{https://github.com/hyperopt/hyperopt} for 50 iterations. The hyper-parameters are given in Table~\ref{tb-sane-hyper-paras}, and we set dropout $p=0.6$ for it performs empirically well. 
	
%
\end{itemize}

\end{document}